\documentclass[table]{article}

\PassOptionsToPackage{authoryear, round}{natbib}

\usepackage[preprint]{neurips_2026}

\usepackage[utf8]{inputenc} 
\usepackage[T1]{fontenc}    
\usepackage{url}            
\usepackage{booktabs}       
\usepackage{amsfonts}       
\usepackage{nicefrac}       
\usepackage{microtype}      
\usepackage[colorlinks,
            linkcolor=red,
            anchorcolor=blue,
            citecolor=green
            ]{hyperref}

\usepackage{xcolor}
\usepackage{natbib}
\setcitestyle{authoryear,round}
\usepackage{graphicx}
\usepackage{amssymb}
\usepackage{multirow}
\usepackage{arydshln}
\usepackage[listings]{tcolorbox}
\tcbuselibrary{breakable}
\usepackage{footmisc}
\usepackage{enumitem}
\usepackage{array}
\usepackage{amsmath}
\usepackage{wrapfig}
\usepackage{needspace}
\usepackage{xspace}
\usepackage{fontawesome5}

\usepackage{algorithm}
\usepackage{algpseudocode}
\usepackage{eso-pic}

\newcommand{\ScienceOneTitleLogoPosFirst}{%
  \AtPageUpperLeft{%
    \hspace{37.7mm}\raisebox{-26.3mm}{\includegraphics[height=9mm]{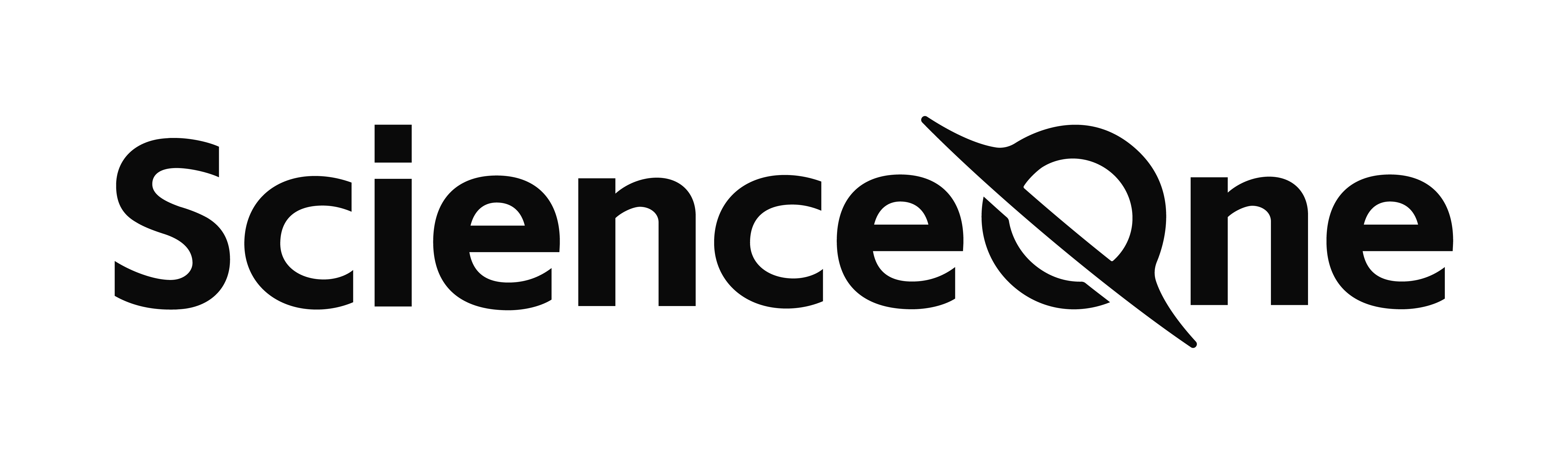}}%
  }%
}
\AddToShipoutPictureFG{%
  \ifnum\value{page}=1\relax
    \ScienceOneTitleLogoPosFirst
  \fi
}

\usepackage{inconsolata}
\usepackage{tabularx}
\usepackage{makecell}
\usepackage{mathrsfs}
\usepackage{amsthm}
\usepackage{subfigure}
\usepackage{caption}
\usepackage{soul}
\usepackage{pifont}
\newcommand{\methodname}{\textsc{SciConsolidate}\xspace}
\definecolor{lightgreen}{rgb}{0.85,1.0,0.85}
\sethlcolor{lightgreen}

\definecolor{+}{RGB}{35, 225, 35}
\definecolor{-}{RGB}{224, 25, 25}

\definecolor{mycyan}{RGB}{0, 158, 150}
\definecolor{algcomment}{RGB}{70, 130, 180}
\algrenewcommand{\algorithmiccomment}[1]{\hfill\textcolor{algcomment}{\(\triangleright\) #1}}
\newtcolorbox{conclusionbox}{
  colback=mycyan!10,
  colframe=white,
  boxrule=0pt,
  arc=4pt,
  left=3pt,
  right=3pt,
  top=3pt,
  bottom=3pt,
}

\newcommand\blfootnote[1]{
    \begingroup
    \renewcommand\thefootnote{}\footnote{#1}
    \addtocounter{footnote}{-1}
    \endgroup
}

\title{From Execution to Capability: Scientific Experience Consolidation via Procedural Knowledge Synthesis}

\author{
\textbf{ScienceOne AI}\\
\textbf{Wenge AI}}

\begin{document}

\maketitle

\begin{center}
\vspace{-6mm}
\small \faGithub\ \href{https://github.com/ScienceOne-AI/SciConsolidate}{\texttt{https://github.com/ScienceOne-AI/SciConsolidate}}
\vspace{3mm}
\end{center}

\begin{abstract}
Large language models increasingly solve scientific-computing tasks, but executable feedback from one problem rarely becomes durable capability on subsequent problems.
We study \emph{scientific-computing experience consolidation}: converting verified runtime experience into transferable procedural knowledge and persistent model improvement.
This setting presents two challenges: trajectory-derived artifacts may encode source-specific repairs rather than cross-task computational mechanisms, and a weaker target model may be unable to operationalize an otherwise valid abstract procedure---an \emph{abstraction--execution gap}.
We introduce \methodname, which contrasts verified successes and failures to induce cross-task procedures, selects them through a development-validation gate, and uses failure-informed, answer-free query synthesis to expand the consolidation data without requiring pre-existing reference answers.
Because the target model may not directly execute these abstractions, a stronger model concretizes them into executable code supervision for standard, procedure-free SFT; a matched no-procedure teacher branch isolates the value of procedural guidance.
On SciCode, runtime procedure injection improves Qwen3.6-27B by +3.85/+6.26 sub-step/main-problem points, but yields almost no aggregate main-problem gain for Qwen3.5-9B, providing operational evidence of the abstraction--execution gap.
After procedure-guided concretization, the 9B student improves under procedure-free deployment by +3.89/+6.25 points over the no-procedure SFT control and by +5.62/+11.25 over the original 9B model.
These results establish an experience-to-capability pathway for scientific computing and provide a practical starting point for scaling self-improving scientific assistance.
\end{abstract}

\section{Introduction}
\label{sec:introduction}

Large language models (LLMs) have recently made substantial progress across scientific tasks~\citep{ma2025scigym,guo2025chemreasoner,gottweis2025aicoscientist,swanson2025aiScientistV2,huang2025biomni}, particularly in scientific computing~\citep{gandhi2025researchcodeagent,hua2025researchcodebench,somasekharan2025cfdllmbench,mohammadzadeh2025fembench}.
Scientific computing, where models help translate scientific specifications into executable numerical, analytical, and simulation procedures, creates a distinctive opportunity for continual improvement because generated procedures can be executed, exposing concrete failures in numerical conventions, array semantics, interfaces, and state transitions.
Current evaluations nevertheless emphasize whether a model solves each problem once, not whether such execution evidence reduces related failures on subsequent problems.
As a result, scientifically informative feedback accumulates in trajectories and logs while the model continues to approach new computations with essentially the same parameters and latent failure modes.
Long-horizon scientific-computing assistance therefore requires a mechanism that abstracts verified execution evidence beyond individual problems, validates its transfer, and returns it to the model as durable capability.
We study this problem as \emph{scientific-computing experience consolidation}: the conversion of verified runtime experience into transferable procedural knowledge and persistent model improvement.

Prior memory and skill systems, developed largely for general reasoning, tool-use, and agent tasks, reuse experience by summarizing trajectories into textual artifacts and refining, retrieving, or coupling them to policy learning~\citep{li2026skillsbench,zhong2026skilllearnbench,wang2025asi,yang2026skillopt,ouyang2025reasoningbank,zhang2026memskill,wu2025evolver,zhang2025memevolve}.
Such artifacts can preserve a successful action or repair for later reuse, but task-level feedback alone does not establish whether they capture a cross-task computational mechanism or merely a source-specific fix.
Scientific computing therefore requires abstraction grounded in executable contrasts between successes and failures, followed by validation beyond the trajectories from which a procedure was induced.
Moreover, dedicated internalization methods train the target policy itself to consume externally provided skills~\citep{lu2026skill0,zhu2026skill05,shi2026skill1,xia2026skillrl,zhang2026skilltolora}, implicitly assuming that it can already extract actionable guidance from them.
Yet a valid abstract procedure is not itself an executable solution: the model must recognize when it applies, bind its clauses to the current scientific state, and sustain the resulting decisions across a multi-step computation.
A weaker target may fail at this instantiation even when the procedure is useful to a stronger model, creating a capacity-dependent \textbf{abstraction--execution gap}.

To address these limitations, we introduce \methodname, which contrasts verified target-model successes and failures to induce cross-task procedures, selects them through a development-validation gate, and uses rollout-derived failure priors to construct balanced scientific-computing queries.
Because the target model may not directly execute the resulting abstractions, a stronger model concretizes them into executable code supervision; the target is then trained by standard SFT and evaluated without runtime procedures, with a matched no-procedure teacher branch isolating the value of procedural guidance.

We evaluate this pathway on SciCode~\citep{tian2024scicode}.
On the aggregate benchmark, runtime procedure injection improves Qwen3.6-27B by 3.85 sub-step points and 6.26 main-problem points, but produces almost no aggregate main-problem gain for Qwen3.5-9B; after procedure-guided concretization and SFT, the 9B student evaluated without runtime procedures improves over the no-procedure SFT control by 3.89/6.25 points and over the original 9B model by 5.62/11.25 points, with a guided-over-control gain of 1.46/3.33 on the frozen held-out split.
These results reveal an \emph{abstraction--execution gap}: useful procedures may be directly operational only for a stronger model, yet their value can be transferred to a weaker model by converting abstract guidance into concrete, executable supervision; the transfer extends beyond the experience-source split but varies across task types and failure families.

Our contributions are threefold:
\begin{itemize}[leftmargin=*, itemindent=0pt]
    \item \textbf{A scientific-computing experience-consolidation formulation.} We formulate the conversion from verified scientific-computing experience to reusable procedures and persistent model capability as a distinct systems problem, and investigate it as a starting point toward self-improving scientific assistance.
    \item \textbf{An empirical characterization of the abstraction--execution gap.} We empirically show that the utility of scientific-computing procedures depends on the consuming model: procedures that materially help a stronger model may not be directly executable by the target model.
    \item \textbf{A procedural knowledge synthesis pathway.} We combine contrastive experience abstraction, validation-gated procedure selection, answer-free scientific-computing query synthesis, and procedure-guided concretization with standard SFT, providing a practical route to scaling consolidation without runtime procedures.
\end{itemize}

\section{Experience-to-Capability Formulation}
\label{sec:preliminary}

\paragraph{Scientific-computing tasks and verified experience.}
Let $\mathcal{D}=\{x_i\}_{i=1}^N$ be scientific-computing tasks drawn from a target distribution $\mathcal{T}$.
Each task $x$ contains an input specification, optional scientific background, and an executable evaluation function $m(\cdot,x)$.
A target model $M_\theta$ produces one or more attempts $r(x)$, and execution yields outcome evidence such as task success, sub-step success, or an observed failure.
We denote the accumulated runtime evidence by
\begin{equation}
\mathcal{E}(M_\theta,\mathcal{D})
=\{(x,r_k(x),m(r_k(x),x))\}_{x\in\mathcal{D},k=1}^{K}.
\end{equation}
Unlike unverified textual traces, this evidence exposes which computational behaviors actually satisfy the scientific task.

\paragraph{Procedural knowledge consolidation.}
From execution evidence, an induction procedure $A$ produces a set of natural-language procedural artifacts $\mathcal{S}=A(\mathcal{E})$.
Each $s\in\mathcal{S}$ targets a family of related failure mechanisms rather than a single problem instance.
We use the term \emph{procedural knowledge} to emphasize three intended properties: it is grounded in observed execution, abstracted across tasks, and selected by transfer utility on a separate development-validation split.
The artifact may be represented as an agent skill or standard operating procedure.
Here it denotes knowledge about reliable scientific computation---for example interface, shape, numerical, or dependency-flow practices---rather than a newly discovered scientific law.
Its utility is determined empirically rather than by textual plausibility alone.

\paragraph{Abstraction--execution gap.}
Let $M(x,s)$ denote inference with procedure $s$ supplied in context.
A validated procedure can be informative while remaining difficult for a target model to operationalize.
We characterize its direct runtime utility for model $M$ as
\begin{equation}
U(M,s)=\mathbb{E}_{x\sim\mathcal{T}}\left[m(M(x,s),x)-m(M(x),x)\right].
\end{equation}
An abstraction--execution gap is present when the same procedure has substantially different utility across consumers, for example $U(M_T,s)>U(M_\theta,s)$ for a stronger model $M_T$ and the target model $M_\theta$.
This definition concerns operational utility, not whether either model can paraphrase the procedure.

\paragraph{Capability-consolidation objective.}
Our deployment goal is an updated model $M_{\theta'}$ that improves without receiving an external procedure:
\begin{equation}
\label{eq:skill_free_objective}
\max_{\theta'}\ \mathbb{E}_{x\sim\mathcal{T}}\left[m(M_{\theta'}(x),x)\right].
\end{equation}
Procedural knowledge may be used during data construction, but not in the evaluated student input.
A stronger model $M_T$ concretizes a selected procedure into a solution $y_i^T=M_T(x_i,s_i)$, yielding
\begin{equation}
\mathcal{D}_{\mathrm{sft}}=\{(x_i,y_i^T)\},
\qquad
\mathcal{L}_{\mathrm{sft}}(\theta)
= - \sum_i \log p_\theta(y_i^T \mid x_i).
\end{equation}
The procedure therefore reaches the student only through executable supervision.
This instantiates one pass through the experience--knowledge--capability pathway:
\begin{equation}
M_\theta
\xrightarrow{\text{execution}} \mathcal{E}
\xrightarrow{\text{abstraction}} \mathcal{S}
\xrightarrow{\text{guided synthesis}} \mathcal{D}_{\mathrm{sft}}
\xrightarrow{\text{learning}} M_{\theta'}.
\end{equation}

\section{Scientific-Computing Experience Consolidation via Procedural Knowledge Synthesis}
\label{sec:method}

\begin{figure*}[t]
    \centering
    \includegraphics[width=\textwidth]{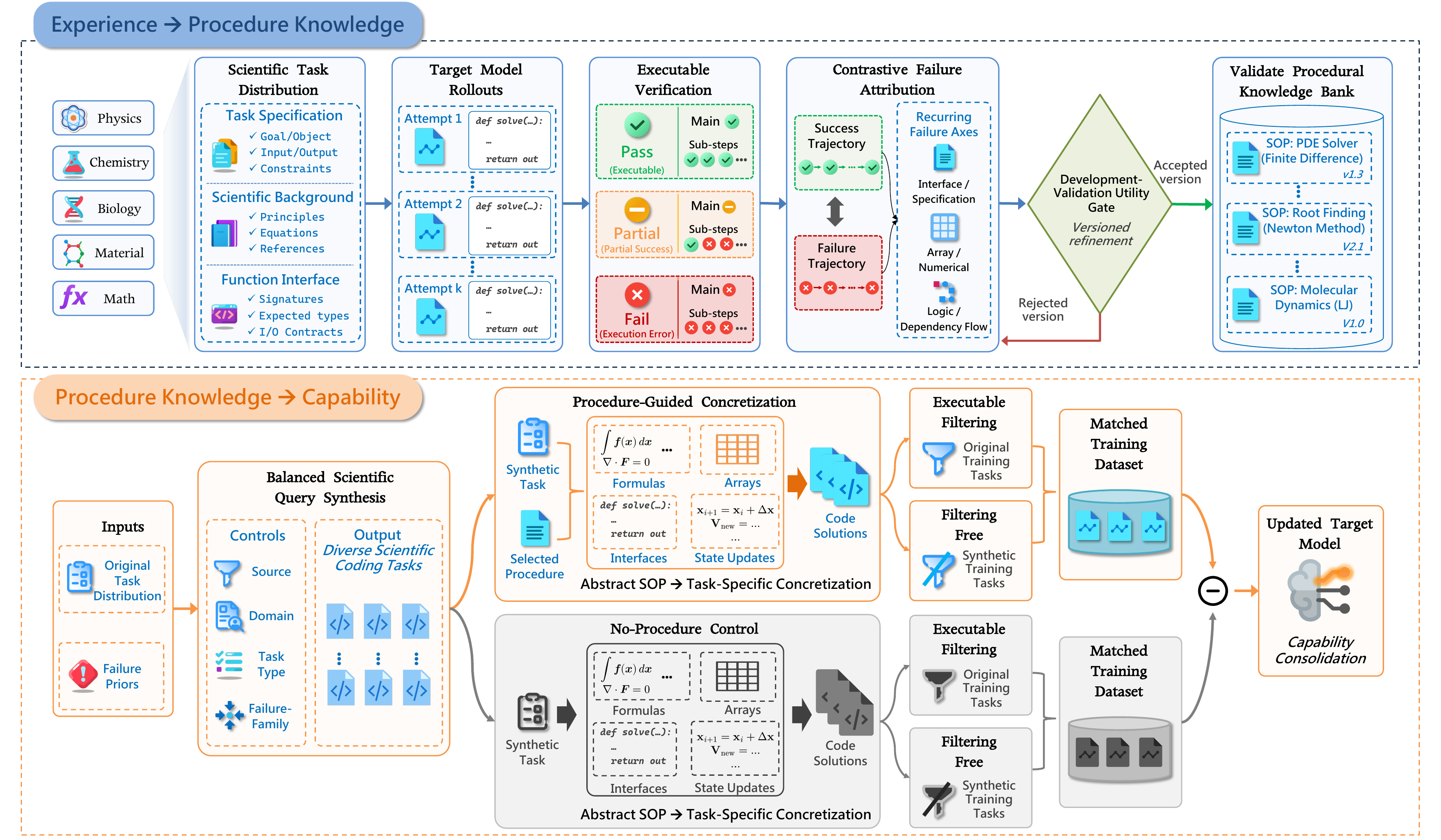}
    \caption{Overview of scientific-computing experience consolidation via procedural knowledge synthesis. The upper path abstracts verified target-model experience into development-validated family-level procedures. The lower path uses those procedures and rollout-derived failure priors to construct scientific-computing queries and executable code supervision, after which the target model is updated and evaluated without runtime procedures. The procedure-guided and no-procedure branches follow the same construction protocol, but their retained sizes can differ after executable filtering. The dashed return denotes a future multi-round extension and is not evaluated in this work; the frozen held-out split is never used for iterative decisions.}
    \label{fig:method_framework}
\end{figure*}

We present \methodname, a data-centric framework for converting verified scientific-computing experience into persistent model capability through procedural knowledge synthesis.
The method does not assume that the target model can directly execute a high-level procedure.
Instead, it uses the target model's own behavior to identify recurring computational failures, synthesizes validation-selected procedures from those failures, expands the experience distribution with balanced scientific-computing query synthesis, and uses a stronger model to instantiate the selected procedures as concrete supervised solutions.
The target model is then trained with standard SFT and evaluated without external procedures.
Figure~\ref{fig:method_framework} illustrates the evaluated experience-to-capability pathway and distinguishes it from the unevaluated multi-round extension.

\subsection{Task Structuring and Split Discipline}
\label{sec:split}

We assign each task a descriptor vector $z(x)$ that captures both scientific and operational structure.
In our instantiation, $z(x)$ includes domain, subdomain, task type, difficulty estimates, and failure-family labels induced from execution evidence.
These descriptors support stratified splitting, cross-task failure discovery, and distribution control during synthetic experience construction.

The task set is partitioned into three disjoint subsets:
\begin{equation}
\mathcal{D}=\mathcal{D}_{\mathrm{train}}\cup \mathcal{D}_{\mathrm{val}}\cup \mathcal{D}_{\mathrm{test}} .
\end{equation}
$\mathcal{D}_{\mathrm{train}}$ supplies target-model attempts, failure analysis, knowledge induction, and training-data seeds.
$\mathcal{D}_{\mathrm{val}}$ is used for development-time selection, including procedure revision, data-mixture selection, and checkpoint selection.
$\mathcal{D}_{\mathrm{test}}$ is frozen and never used for iterative decisions.
This discipline is essential because both natural-language procedures and synthetic data mixtures are optimized artifacts; without a development gate, either can encode source-specific regularities that resemble knowledge but fail to transfer.

\subsection{Verified Experience Collection and Failure Attribution}
\label{sec:skill_induction}

We run the target model $M_\theta$ on $\mathcal{D}_{\mathrm{train}}$ and collect multiple attempts
$R(x)=\{r_1(x),\ldots,r_K(x)\}$ for each task.
Executable evaluation separates successful, partially successful, and failed behavior at the task and sub-step levels.
Multiple attempts are important because they provide within-task contrasts: the same model can expose both a valid computational path and a nearby failure under nearly identical task semantics.

We compare successful and failed attempts to identify recurring mechanisms rather than grouping errors only by scientific domain.
For example, shape misalignment can appear in physics simulation and computational chemistry; interface misuse can affect numerical methods and biological modeling; and broken dependency flow can occur wherever a multi-step computation updates live state.
This produces failure families $\{F_l\}_{l=1}^{L}$ whose members share an operational failure axis.

For each family $F_l$, we induce an initial procedure $s_l^{(0)}$ from paired execution evidence.
The procedure describes a reusable intervention---activation conditions, decision rules, termination conditions, and high-risk anti-patterns---rather than a solution template for an individual scientific task.
Grounding the abstraction in observed contrasts helps distinguish a procedure that explains \emph{why behavior succeeds} from a summary that merely restates \emph{what a successful solution contains}.

\subsection{Validation-Gated Procedural Knowledge Synthesis}
\label{sec:skill_optimization}

The induced procedures are optimized as versioned knowledge artifacts.
For family $F_l$, let $\{s_l^{(t)}\}_{t\ge0}$ denote successive revisions.
At iteration $t$, additional target-model evidence from $\mathcal{D}_{\mathrm{train}}$ is used to propose $\tilde{s}_l^{(t+1)}$.
The candidate is promoted only if it improves a paired utility gate on $\mathcal{D}_{\mathrm{val}}$:
\begin{equation}
s_l^{(t+1)} =
\begin{cases}
\tilde{s}_l^{(t+1)}, & \text{if } G(\mathcal{D}_{\mathrm{val}}, \tilde{s}_l^{(t+1)}) > G(\mathcal{D}_{\mathrm{val}}, s_l^{(t)}),\\
s_l^{(t)}, & \text{otherwise.}
\end{cases}
\end{equation}
Here $G$ measures empirical procedure utility, such as paired improvement over the previous revision or a no-procedure baseline.
The gate treats transfer as part of the operational definition of reusable procedural knowledge: a plausible rule is not promoted merely because it explains its source failures.
The selected artifact $s_l^\star$ is the best development-validated version for family $F_l$.

\subsection{Knowledge-Guided Scientific-Computing Query Synthesis}
\label{sec:balanced_synthesis}

The original task set is too small and pattern-concentrated to expose a procedure across diverse scientific contexts.
We therefore expand the experience pool with additional queries generated under descriptor $z(x)$ and the validated failure structure:
\begin{equation}
\mathcal{Q}_{\mathrm{syn}}
= \operatorname{Balance}\!\left(
\operatorname{Generate}(\mathcal{D}_{\mathrm{train}},z,\{s_l^\star\}_{l=1}^{L})
\right).
\end{equation}
The generator produces new problem specifications, function interfaces, dependencies, and candidate executable tests while preserving the broad scientific-computing character of the seeds.
Pre-existing benchmark answers are not required: concrete supervision is supplied in the next stage, and executable checks are used as filters when available.

Balancing is central to the experience-consolidation objective.
Uncontrolled generation can overproduce tasks from a few source problems, domains, or failure families, causing the learned model to memorize a narrow realization of the procedure.
We therefore control source-problem concentration, domain and task-type shares, failure-family coverage, and replay of selected target-model behaviors.
The resulting training-query pool is
\begin{equation}
\mathcal{Q}_{\mathrm{sft}}=\mathcal{Q}_{\mathrm{orig}}\cup\mathcal{Q}_{\mathrm{syn}}.
\end{equation}

\subsection{Procedure-Guided Knowledge Concretization}
\label{sec:teacher_concretization}

A validated procedure may still lie across the abstraction--execution gap of the target model.
We therefore use a stronger model $M_T$ as a temporary \emph{procedure concretizer}.
For query $x$ assigned to family $F_l$, the concretizer receives
\begin{equation}
q_T(x)=\operatorname{concat}(x,s_l^\star)
\end{equation}
and produces a concrete scientific-computing solution
\begin{equation}
y^T=M_T(q_T(x)).
\end{equation}
The student input remains procedure-free, $q_S(x)=x$, so the training example is $(x,y^T)$.
The concretizer performs the difficult binding from abstract decision rules to task-specific formulas, arrays, interfaces, and state transitions; the student learns the resulting executable behavior without parsing the procedural artifact at deployment time.

For original development-train tasks, only verifier-passing solutions are retained.
For synthesized queries, concretizer-generated solutions do not require pre-existing benchmark gold answers; locally available executable checks are used for filtering.
In the current implementation, the supervised target is the final code answer rather than the concretizer's hidden reasoning trace, reducing direct leakage of privileged procedure text into the student's output.

To probe the value of procedural guidance beyond generic capability transfer from the 27B model, we construct a no-procedure control using the same query-construction and SFT pipeline:
\begin{equation}
\mathcal{D}_{\mathrm{sft}}^0=\{(x_i,y_i^0)\},
\qquad y_i^0=M_T(x_i).
\end{equation}
The guided student is trained on $\mathcal{D}_{\mathrm{sft}}=\{(x_i,y_i^T)\}$ and the control student on $\mathcal{D}_{\mathrm{sft}}^0$, both with procedure-free inputs.
Their comparison asks whether validated procedural knowledge changes the supervision in a way that survives after the external artifact is removed.
This capability-consolidation step completes the evaluated experience-to-capability pathway.
Repeating all stages with the updated model is a natural route toward multi-round scientific self-improvement, but is outside the scope of the present study.

\section{Experiments}
\label{sec:experiments}

In this section, our empirical experiments aim to answer two key questions:
\begin{itemize}[leftmargin=*, itemsep=0pt, topsep=2pt]
    \item \emph{Q1: Does procedural knowledge synthesized from verified execution exhibit operational utility beyond its source tasks, and how does that utility depend on the consuming model?}
    \item \emph{Q2: Can such knowledge be converted into persistent target-model capability through procedure-guided supervision synthesis?}
\end{itemize}

\subsection{Experimental Setup}
\label{sec:exp_setup}

\paragraph{Evaluation splits and metrics.}
We evaluate the proposed framework in SciCode~\citep{tian2024scicode}, whose scientific-computing problems, deterministic execution, decomposed sub-problems, and explicit constraints on formulas, arrays, units, interfaces, and state transitions make experience attributable at fine granularity.
This structure makes the benchmark suitable for tracing the path from observed failures to procedural knowledge and model updates.
The evaluation uses 80 main problems with 338 decomposed sub-steps.
We report sub-step accuracy and main-problem accuracy.
The evaluation is organized into three named parts.
The \emph{development-train} split contains 35 main problems and 152 sub-steps; it serves as the rollout/source pool for skill discovery and development decisions.
The \emph{development-valid} split contains 15 main problems and 46 sub-steps; it is held out from the source pool but used for procedure, mixture, and checkpoint selection.
The \emph{held-out} split contains 30 main problems and 140 sub-steps; it is kept frozen and used only for generality evaluation.
We additionally report aggregate performance over all 80 problems, which combines development-train, development-valid, and held-out.
Because development-valid participates in model development, only the frozen held-out split should be interpreted as an untouched generalization evaluation; the aggregate is a compact summary rather than an independent test estimate.

\paragraph{Models and conditions.}
We evaluate four model groups:
(i) the Qwen3.5-9B base model,
(ii) the Qwen3.6-27B concretizer model,
(iii) a 9B student trained on procedure-guided 27B solutions,
and (iv) a pipeline-matched 9B student trained on no-procedure 27B solutions.
Both students are obtained by full-parameter SFT under the same optimization configuration; the complete training hyperparameters are reported in Appendix~\ref{app:training_hyperparameters}.
For the 9B base model and 27B concretizer, we compare inference with and without the runtime procedure bundle.
For student models, the key deployment condition is background-conditioned inference without runtime procedure injection.
Unless otherwise stated, the fine-grained diagnostics below are computed from the same main-result-aligned runs as the main result tables.
We use GLM-5.1~\citep{zai2026glm51} to label failure modes and families, and then distill and iteratively refine the procedural artifacts used in these experiments. The specific prompts are provided in Appendix~\ref{app:skill_distillation_prompts}.

\paragraph{Synthetic supervision construction.}
The SFT data is built from development-train rollouts and synthetic tasks derived from development-train priors.
In the main procedure-guided run, the retained training pool contains 2,095 examples: 395 examples from the original development-train task distribution and 1,700 synthetic examples generated from balanced query expansion.
The no-procedure control uses the same source pool, candidate-generation budget, and construction protocol, retaining 1,985 examples after executable filtering.
Target-model rollouts provide task-family and failure-mode priors for balanced query synthesis; the 27B concretizer then supplies executable code supervision under either procedure-guided or no-procedure prompting.
Executable checking filters generated solutions when local tests are available; the synthetic tasks do not require pre-existing benchmark reference answers.
Additional data-composition statistics are provided in Appendix~\ref{app:exp_diagnostics}.

\subsection{The Abstraction--Execution Gap}
\label{sec:capacity_mismatch}

We first ask whether validated procedural knowledge is equally executable by models of different capacities.
We compare the 9B target model and the 27B model under the same background-conditioned setting, with and without the same runtime procedure bundle.

\begin{table}[t]
\centering
\small
\caption{Direct runtime use of the induced procedural knowledge under the background-conditioned setting. Results are sub-step/main-problem accuracy (\%).}
\label{tab:capacity_mismatch}
\resizebox{\linewidth}{!}{
\begin{tabular}{llcccc}
\toprule
\textbf{Model} & \textbf{Runtime procedure} & \textbf{development-train} & \textbf{development-valid} & \textbf{held-out} & \textbf{all} \\
\midrule
\multirow{3}{*}{Qwen3.5-9B base} & No & 28.95 / 11.43 & 34.78 / 13.33 & 29.29 / 3.33 & 29.88 / 8.75 \\
 & Yes & 30.26 / 14.29 & 34.78 / 0.00 & 27.14 / 6.67 & 29.58 / 8.75 \\
 & Gain & +1.31 / +2.86 & +0.00 / -13.33 & -2.15 / +3.34 & -0.30 / +0.00 \\
\midrule
\multirow{3}{*}{Qwen3.6-27B concretizer} & No & 45.39 / 22.86 & 52.17 / 20.00 & 36.43 / 3.30 & 42.60 / 14.99 \\
 & Yes & 50.66 / 28.57 & 45.65 / 20.00 & 42.14 / 13.33 & 46.45 / 21.25 \\
 & Gain & +5.27 / +5.71 & -6.52 / +0.00 & +5.71 / +10.03 & +3.85 / +6.26 \\
\bottomrule
\end{tabular}
}
\end{table}

Table~\ref{tab:capacity_mismatch} provides operational evidence of an abstraction--execution gap.
The 27B model gains 3.85 sub-step points and 6.26 main-problem points from the runtime procedures, while the 9B model obtains no corresponding main-accuracy gain.
Because the same artifacts benefit the stronger consumer, their failure on the 9B model cannot be attributed only to an absence of useful information.
Instead, the result is consistent with a consumer-dependent operationalization barrier between an abstract procedure and its concrete scientific execution.
The split-level results are consistent with this interpretation: on development-train, the 27B concretizer improves by 5.27/5.71 sub-step/main points after procedure injection, whereas the 9B base model moves only 1.31/2.86.
On the frozen held-out split, the 27B model's main accuracy increases from 3.30 to 13.33, while the 9B model remains much less responsive.
The negative development-valid sub-step delta also shows that procedure utility is not uniform across task subsets; our claim is consumer-dependent operational value, not universal improvement from every procedure injection.

We further decompose the same runs by SciCode task metadata.
These diagnostics report sub-step accuracy because the per-domain main-problem counts are small.
The decomposition in Table~\ref{tab:capacity_breakdown} shows that the 27B gain is not a uniform global shift: it is strongest on scientific-calculation steps and appears primarily in Biology, Chemistry, and Physics.
In contrast, the 9B response to the same runtime procedures is mixed across buckets and does not produce a consistent aggregate gain.
This supports the abstraction--execution interpretation: validated procedural knowledge can be useful, but a smaller model does not reliably convert the high-level instruction into executable scientific behavior.
Appendix~\ref{app:direct_family_skill} provides an additional development-side diagnostic over labeled failure families and a representative capacity-mismatch case.

\begin{table}[t]
\centering
\small
\caption{Fine-grained direct procedure use. Each cell reports sub-step accuracy without runtime procedures $\rightarrow$ with runtime procedures, followed by the gain in parentheses.}
\label{tab:capacity_breakdown}
\begingroup
\setlength{\tabcolsep}{4pt}
\begin{tabular*}{\linewidth}{@{\extracolsep{\fill}}llccc@{}}
\toprule
\textbf{Axis} & \textbf{Bucket} & \textbf{\# steps} & \textbf{Qwen3.5-9B base} & \textbf{Qwen3.6-27B concretizer} \\
\midrule
\multirow{3}{*}{Task type} & Numerical method & 65 & 24.6 $\rightarrow$ 26.2 (+1.5) & 40.0 $\rightarrow$ 38.5 (-1.5) \\
& Scientific calculation & 167 & 28.1 $\rightarrow$ 26.9 (-1.2) & 39.5 $\rightarrow$ 47.3 (+7.8) \\
& Simulation & 106 & 35.8 $\rightarrow$ 35.8 (+0.0) & 49.1 $\rightarrow$ 50.0 (+0.9) \\
\midrule
\multirow{5}{*}{Domain} & Biology & 27 & 25.9 $\rightarrow$ 18.5 (-7.4) & 48.1 $\rightarrow$ 55.6 (+7.4) \\
& Chemistry & 53 & 20.8 $\rightarrow$ 24.5 (+3.8) & 35.8 $\rightarrow$ 41.5 (+5.7) \\
& Material Science & 58 & 48.3 $\rightarrow$ 51.7 (+3.4) & 48.3 $\rightarrow$ 46.6 (-1.7) \\
& Math & 33 & 33.3 $\rightarrow$ 30.3 (-3.0) & 45.5 $\rightarrow$ 42.4 (-3.0) \\
& Physics & 167 & 26.3 $\rightarrow$ 25.1 (-1.2) & 41.3 $\rightarrow$ 47.3 (+6.0) \\
\bottomrule
\end{tabular*}
\endgroup
\end{table}

\subsection{Returning Procedural Knowledge to Model Capability}
\label{sec:skill_internalization_results}

We next ask whether procedural knowledge can improve the 9B model after the external artifact is removed.
We focus on background-conditioned deployment without runtime procedure injection.
This tests the knowledge-to-capability transformation: whether procedure-guided solutions alter the student's endogenous behavior beyond generic supervision from the same stronger model.

\subsubsection{Aggregate Procedure-Free Deployment}

\begin{table}[t]
\centering
\small
\caption{Procedure-free deployment under the background-conditioned setting. Results are sub-step/main-problem accuracy (\%).}
\label{tab:sft_main}
\resizebox{\linewidth}{!}{
\begin{tabular}{lcccc}
\toprule
\textbf{Model} & \textbf{development-train} & \textbf{development-valid} & \textbf{held-out} & \textbf{all} \\
\midrule
9B base & 28.95 / 11.43 & 34.78 / 13.33 & 29.29 / 3.33 & 29.88 / 8.75 \\
No-procedure SFT & 35.53 / 20.00 & 34.78 / 13.33 & 26.40 / 6.67 & 31.61 / 13.75 \\
Procedure-guided SFT & 39.47 / 25.71 & 45.65 / 26.67 & 27.86 / 10.00 & 35.50 / 20.00 \\
\midrule
Gain over no-procedure SFT & +3.94 / +5.71 & +10.87 / +13.34 & +1.46 / +3.33 & +3.89 / +6.25 \\
Gain over 9B base & +10.52 / +14.28 & +10.87 / +13.34 & -1.43 / +6.67 & +5.62 / +11.25 \\
\bottomrule
\end{tabular}
}
\end{table}

Table~\ref{tab:sft_main} compares the three relevant 9B conditions.
The no-procedure SFT model controls for generic transfer from stronger-model-generated code, while the procedure-guided SFT condition uses the consolidated procedures during solution generation.
The two retained datasets are close but not identical in size because executable filtering preserves 2,095 guided examples and 1,985 control examples; this yield difference is part of the practical pipeline, and also prevents interpreting the comparison as a perfectly size-matched causal estimate.
Under procedure-free inference, the guided student improves over the no-procedure SFT baseline by 3.89 sub-step points and 6.25 main-problem points on the aggregate 80-problem summary.
It also improves over the original 9B base model by 5.62 sub-step points and 11.25 main-problem points in the same deployment condition.
Because the external procedure is absent at evaluation time, these gains provide evidence that procedure-guided execution experience has been consolidated into the student's behavior.
The split-level view shows that the gain is not confined to development-train: the guided student also improves over the no-procedure SFT baseline on development-valid by 10.87/13.34 and on the frozen held-out split by 1.46/3.33.
The development-valid result is development evidence because that split participates in selection; the held-out result is the relevant generalization check and is positive but modest.
Together, the results support reusable behavior transfer while also showing that most of the measured improvement is concentrated on the development distribution.
Appendix~\ref{app:exp_diagnostics} provides complementary data, runtime-procedure, and failure-family diagnostics for the same comparison.

\begin{figure}[t]
\centering
\includegraphics[width=\linewidth]{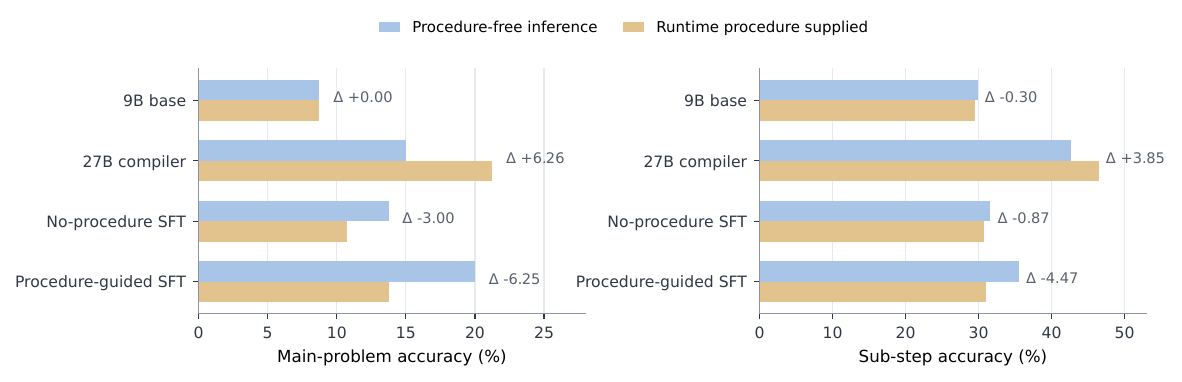}
\caption{Runtime-procedure effects under the background-conditioned aggregate evaluation. The left panel reports main-problem accuracy and the right panel reports sub-step accuracy. Both panels compare procedure-free inference with runtime procedure injection across the four model groups.}
\label{fig:main_effects}
\end{figure}

\subsubsection{Task-Type and Scientific-Domain Breakdown}

Table~\ref{tab:internalization_breakdown} breaks down the aggregate procedure-free student deployment condition by task type and domain.
The guided student improves over the no-procedure SFT control on scientific-calculation and simulation steps, with the strongest task-type gain on simulation.
At the domain level, the gain is positive in Chemistry, Material Science, Math, and Physics, and tied in Biology.
The only negative bucket relative to the no-procedure SFT control is numerical method, where the guided student still remains above the original 9B base model.
This indicates that the aggregate capability-consolidation signal is not confined to a single scientific domain, while also identifying numerical methods as a remaining bottleneck.
Because these buckets pool all splits, they diagnose where the observed gains occur but should not be read as independent domain-level generalization estimates.

\begin{table}[t]
\centering
\small
\caption{Fine-grained procedure-free student deployment. Results are sub-step accuracy (\%). The gain column compares Procedure-guided SFT against No-procedure SFT.}
\label{tab:internalization_breakdown}
\begingroup
\setlength{\tabcolsep}{3pt}
\begin{tabular*}{\linewidth}{@{\extracolsep{\fill}}llccccc@{}}
\toprule
\textbf{Axis} & \textbf{Bucket} & \textbf{\# steps} & \textbf{9B base} & \makecell{\textbf{No-procedure}\\\textbf{SFT}} & \makecell{\textbf{Procedure-guided}\\\textbf{SFT}} & \textbf{Gain} \\
\midrule
\multirow{3}{*}{Task type} & Numerical method & 65 & 24.6 & 32.3 & 29.2 & -3.1 \\
& Scientific calculation & 167 & 28.1 & 28.7 & 31.7 & +3.0 \\
& Simulation & 106 & 35.8 & 35.8 & 45.3 & +9.4 \\
\midrule
\multirow{5}{*}{Domain} & Biology & 27 & 25.9 & 33.3 & 33.3 & +0.0 \\
& Chemistry & 53 & 20.8 & 17.0 & 22.6 & +5.7 \\
& Material Science & 58 & 48.3 & 44.8 & 53.4 & +8.6 \\
& Math & 33 & 33.3 & 36.4 & 42.4 & +6.1 \\
& Physics & 167 & 26.3 & 30.5 & 32.3 & +1.8 \\
\bottomrule
\end{tabular*}
\endgroup
\end{table}

\subsubsection{Failure-Family Diagnostics}

Finally, we examine the failure-family structure used by our skill-discovery pipeline.
These labels are available for development-train failure and partial-success steps, where target-model rollouts were clustered into failure axes before procedure synthesis; unlabeled steps are omitted from this diagnostic.

Table~\ref{tab:failure_family_breakdown} shows that procedure-guided SFT improves over no-procedure SFT on the array/numerical family and the logic/dependency-flow family.
The array/numerical family is positive in this labeled failure-mode view, even though the broader numerical-method task type remains negative above.
This difference is expected: F2 covers a narrower set of shape, dtype, and numerical-stability failures, whereas the numerical-method task type also includes algorithmic and formula-heavy steps that are not fully addressed by the current procedures.
The specification/interface family is slightly negative in this diagnostic, indicating that interface-auditing behavior is not yet transferred as stably as the aggregate task-type gains might suggest.

Overall, the pattern suggests that the evaluated pathway returns several failure-axis procedures to model behavior, while broader numerical-method reasoning still requires more targeted experience and supervision.
Appendix~\ref{app:exp_diagnostics} further reports the pipeline-matched data statistics, runtime-procedure deltas, and representative rescued cases used to interpret these diagnostics.

\begin{table}[t]
\centering
\small
\caption{Development-train diagnostic by learned failure family. Results are sub-step accuracy (\%) over labeled failure/partial-success steps jointly covered by the compared runs.}
\label{tab:failure_family_breakdown}
\begingroup
\setlength{\tabcolsep}{3pt}
\begin{tabular*}{\linewidth}{@{\extracolsep{\fill}}llccccc@{}}
\toprule
\textbf{Family} & \textbf{Failure axis} & \textbf{\# steps} & \textbf{9B base} & \makecell{\textbf{No-procedure}\\\textbf{SFT}} & \makecell{\textbf{Procedure-guided}\\\textbf{SFT}} & \textbf{Gain} \\
\midrule
F1 & Specification/interface & 29 & 12.1 & 18.1 & 15.5 & -2.6 \\
F2 & Array/numerical reasoning & 14 & 28.6 & 17.9 & 37.5 & +19.6 \\
F3 & Logic/dependency flow & 11 & 43.2 & 34.1 & 40.9 & +6.8 \\
F4 & Knowledge/capacity gap & 59 & 11.0 & 11.9 & 15.7 & +3.8 \\
\bottomrule
\end{tabular*}
\endgroup
\end{table}

\subsection{Reintroducing External Procedures After Consolidation}
\label{sec:runtime_skill_after_sft}

The deployment-oriented comparison above evaluates the students without runtime procedures, which is also their strongest setting in Figure~\ref{fig:main_effects}.
This contrasts with the 27B model, for which the same procedures are useful at inference time.
Runtime procedure prompting is therefore not simply an additive information channel for the 9B model; it also changes the input distribution and imposes an additional planning interface.
Before SFT, the 9B model mostly fails to operationalize the abstraction.
After SFT, it has learned a direct mapping from ordinary scientific task inputs to concretizer-generated code, so reintroducing the high-level artifact can become redundant or disruptive.
The larger degradation for the guided student is consistent with---but does not uniquely prove---the interpretation that the relevant behavior has already been compiled into the training solutions.
Operationally, the result favors using external procedures as data-construction conditions rather than permanent runtime dependencies for the smaller deployed model.

\subsection{Interpretation}
\label{sec:interpretation}

The results support two linked claims about the experience-to-capability pathway.
First, reusable procedural knowledge and executable capability are distinct: the stronger 27B model benefits from runtime procedures, while the 9B model does not reliably operationalize them.
Second, this abstraction--execution gap can be bridged during data construction.
By letting the stronger model consume the validated procedures and training the 9B model on the resulting concrete solutions, the target improves under procedure-free inference over both its initial state and a no-procedure SFT control.
This supports procedure-guided supervision synthesis as a practical route from consolidated experience to model capability.
The fine-grained diagnostics add two qualifications.
The positive effect is broad enough to appear across several task types, domains, and failure families, but it is not uniform: broader numerical-method reasoning and specification/interface transfer remain weak buckets.
These buckets identify where a subsequent loop should collect and synthesize more targeted experience, especially formula-intensive numerical methods and specification-sensitive interfaces.
The present experiments establish one pass through the loop rather than demonstrating autonomous multi-round improvement.

\section{Related Work}
\label{sec:related_work}

\paragraph{Scientific tasks and scientific computing.}
Recent work has extended LLM reasoning and agentic systems across scientific problem solving, chemistry, biomedicine, hypothesis generation, and increasingly autonomous research workflows~\citep{ma2025scigym,guo2025chemreasoner,gottweis2025aicoscientist,swanson2025aiScientistV2,huang2025biomni}.
Within this broader landscape, scientific-computing systems translate scientific specifications into executable research code, numerical analyses, and simulations~\citep{gandhi2025researchcodeagent,hua2025researchcodebench,somasekharan2025cfdllmbench,mohammadzadeh2025fembench}.
These tasks offer more than final-answer supervision: execution reveals whether formulas, interfaces, array semantics, numerical procedures, and state transitions are operationally correct.
SciCode~\citep{tian2024scicode} provides the testbed for our study through scientist-curated problems, decomposed sub-steps, and deterministic tests. We use this verifiable setting to ask whether repeated execution successes and failures can be abstracted, validated across tasks, and returned to the model as persistent capability.

\paragraph{Experience reuse and self-evolving agents.}
A broad line of work seeks to preserve agent experience beyond a single interaction.
Dynamic Cheatsheet~\citep{suzgun2025dynamiccheatsheet}, ReasoningBank~\citep{ouyang2025reasoningbank}, MemP~\citep{fang2025memp}, MemSkill~\citep{zhang2026memskill}, and MemRL~\citep{zhang2026memrl} study adaptive memories, reusable reasoning, procedural-memory construction, memory-management skills, and runtime memory optimization.
EvolveR~\citep{wu2025evolver}, MemEvolve~\citep{zhang2025memevolve}, AutoRefine~\citep{qiu2026autorefine}, and RSEA~\citep{nguyen2026rsea} broaden this view toward experience-driven or recursively selected self-improvement.
These systems motivate our experience--knowledge perspective: runtime histories should become persistent assets rather than isolated traces.
Our study isolates one executable experience-to-capability pathway in scientific computing---verified experience, cross-task abstraction, validation, guided data synthesis, and parameter update---rather than claiming a fully autonomous or indefinitely repeated evolution process.

\paragraph{Agent skills as procedural knowledge artifacts.}
Agent skills provide a concrete representation for consolidated experience.
SkillsBench~\citep{li2026skillsbench}, SkillLearnBench~\citep{zhong2026skilllearnbench}, ASI~\citep{wang2025asi}, AgentSkillOS~\citep{li2026agentskillos}, and SkillLens~\citep{huang2026skilllens} study skill creation, evaluation, organization, and consumer-dependent utility.
SkillOpt~\citep{yang2026skillopt}, related optimization methods~\citep{wang2026skillgrad,gong2026skillmoo}, and SkillEvolver~\citep{zhang2026skillevolver} treat skill text or skill packages as objects that can be iteratively edited and validated.
Governance-oriented systems further address provenance, routing, lifecycle management, and safety~\citep{liu2026skillsvote,huang2026skillwiki,zhu2026skillcoach,pan2026skillguard,zhuang2026agenttrap,jia2026skillsislands}.
We use compact skills or SOPs as the representation of family-level procedural knowledge, but our primary object of study is the conversion chain around them: execution evidence grounds the artifact, development-validation utility determines whether it is retained, and procedure-guided concretization determines whether it can become executable student behavior.

\paragraph{Skill internalization and experience-driven policy learning.}
The closest work studies whether external procedures can be absorbed into model parameters.
SKILL0~\citep{lu2026skill0} uses skills as training-time scaffolds and progressively removes them through reinforcement learning; Skill0.5~\citep{zhu2026skill05} distinguishes general skills to internalize from task-specific skills that may remain externally routed.
Skill1~\citep{shi2026skill1} couples skill selection, utilization, and distillation under outcome feedback; SkillRL~\citep{xia2026skillrl} recursively co-evolves skill banks and policies; OPD-Evolver~\citep{zhang2026opdevolver} distills experience-lifecycle decisions into a deployable policy; Evolving-RL~\citep{fan2026evolvingrl} evaluates both skill-augmented and no-skill behavior; and Skill-to-LoRA~\citep{zhang2026skilltolora} replaces skill text with skill-specific adapters at runtime.

Our work shares the goal of reducing deployment-time dependence on external artifacts, but differs in both its starting point and scope.
Prior internalization methods largely begin with an existing skill representation and optimize a path from given skills to a skill-free policy; we instead begin with the target model's verified scientific-computing experience, including successful and failed code, sub-step execution outcomes, and recurring computational errors.
The resulting pathway is correspondingly broader: verified execution experience is abstracted into cross-task failure mechanisms, promoted into procedures only after development validation, concretized as executable supervision, and finally consolidated into persistent target-model capability.
This also raises a distinct upstream question: how can verified execution history become supervision when the target model cannot reliably operationalize the abstraction it helped reveal?
We answer by moving abstract procedure use to a stronger temporary concretizer during data synthesis, while retaining standard, procedure-free SFT for the target model.
Internalization is therefore one stage of a broader scientific-computing experience-consolidation pathway rather than the sole object of the framework.

\section{Conclusion and Limitations}
\label{sec:conclusion}

\paragraph{Conclusion.}
We studied scientific-computing experience consolidation: how verified scientific-computing experience can be converted into reusable procedural knowledge and then into improved model capability.
Our framework collects multiple target-model attempts, contrasts successful and failed computations, synthesizes validation-selected family procedures, expands the experience distribution through answer-free scientific-computing query synthesis, and uses a stronger model to concretize abstract procedures into code supervision.
SciCode reveals an abstraction--execution gap: procedures that materially help the 27B model do not directly improve the 9B model's main-problem accuracy.
Yet when those procedures guide 27B solution generation, the resulting 9B student improves under procedure-free deployment over both its initial state and a no-procedure SFT control.
These results show that useful scientific-computing experience need not remain a runtime prompt or memory entry; it can be compressed, validated, re-instantiated across new tasks, and partially consolidated into model behavior through standard SFT.
We view this as an evaluated experience-to-capability pathway and a potential building block for future closed-loop improvement, rather than a complete realization of autonomous scientific self-evolution.

\paragraph{Limitations.}
Several limitations define the boundary of the present claims:
\begin{itemize}[leftmargin=*]
    \item \textbf{One scientific testbed and one pathway evaluation.} The evidence is based on SciCode and does not establish generality across scientific environments, model families, long-horizon tool use, or physical experimentation. We also do not repeat the full process with the updated student, so the study demonstrates experience consolidation rather than sustained continual self-improvement.
    \item \textbf{Operational rather than mechanistic evidence for the abstraction--execution gap.} The 9B/27B contrast is consistent with a capacity-dependent ability to instantiate abstract procedures, but two differently pretrained model sizes do not isolate the causal mechanism. A broader scaling study and controlled procedure-complexity analysis are needed.
    \item \textbf{Procedural knowledge is narrower than scientific discovery.} The induced artifacts capture computational practices such as interface discipline, array reasoning, numerical checks, and dependency flow. They do not constitute newly discovered scientific laws, autonomous hypothesis generation, or a closed experimental-science cycle.
    \item \textbf{Dependence on stronger external models.} Failure labeling, procedure induction, and solution generation use models stronger than the 9B target. A fully autonomous self-improving system would need these roles to converge toward the improving model itself while preserving independent validation.
    \item \textbf{Validation of synthesized experience.} Synthetic queries do not require pre-existing reference answers, but their generated solutions are only as reliable as the available executable filters. The current results are strongest where deterministic evaluation is available; less verifiable scientific-computing tasks require richer verification, uncertainty estimation, and recovery mechanisms.
    \item \textbf{Final-answer supervision.} Privileged procedure text can leak into long reasoning traces. We therefore train on final code answers rather than hidden concretizer reasoning, which reduces direct textual leakage but may discard useful deliberative structure and leaves reasoning-preserving consolidation unresolved.
    \item \textbf{Non-uniform transfer and possible negative interference.} Gains are strongest for simulation and several array or dependency-flow failures, while broader numerical methods and specification/interface behavior remain weak. Subsequent loop iterations must target these residual failures without degrading already consolidated behavior.
\end{itemize}

\section{Author Contributions}
\textbf{Core Contributors:} Liwei Dong\textsuperscript{*}, Jiahao Zhao, Nan Xu.
\blfootnote{\textsuperscript{*} Corresponding authors: \texttt{liwei.dong@wenge.com}, \texttt{viviankeith.vk@gmail.com}}

\newpage
\bibliographystyle{unsrtnat}
\bibliography{custom}

\newpage
\appendix

\section{Additional Method Details}
\label{app:method_details}

\definecolor{apppromptbg}{RGB}{246, 248, 252}
\definecolor{apppromptborder}{RGB}{140, 155, 180}
\definecolor{apppromptrole}{RGB}{128, 82, 45}
\definecolor{appskillbg}{RGB}{240, 244, 250}
\definecolor{appskillborder}{RGB}{120, 140, 170}
\definecolor{appcasebg}{RGB}{245, 248, 252}
\definecolor{appcaseborder}{RGB}{62, 95, 135}
\definecolor{appcasetitle}{RGB}{46, 88, 135}
\newcommand{\appPass}{\textcolor{green!45!black}{\ding{51} pass}}
\newcommand{\appFail}{\textcolor{red!65!black}{\ding{55} fail}}
\newcommand{\appPromptRole}[1]{\textcolor{apppromptrole}{{\small\bfseries\ttfamily #1}}}
\tcbset{
  appendixpromptbox/.style={
    colback=apppromptbg,
    colframe=apppromptborder,
    fonttitle=\bfseries\small,
    boxrule=0.6pt,
    arc=2mm,
    left=4pt,
    right=4pt,
    top=3pt,
    bottom=3pt,
    breakable,
    listing only,
    listing options={basicstyle=\ttfamily\footnotesize, breaklines=true, columns=fullflexible, keepspaces=true, escapeinside={(*@}{@*)}},
  },
  appendixskillbox/.style={
    colback=appskillbg,
    colframe=appskillborder,
    fonttitle=\bfseries\small,
    boxrule=0.5pt,
    arc=1.5mm,
    left=4pt,
    right=4pt,
    top=3pt,
    bottom=3pt,
    breakable,
    listing only,
    listing options={basicstyle=\ttfamily\footnotesize, breaklines=true, columns=fullflexible, keepspaces=true},
  },
  appendixcasebox/.style={
    colback=appcasebg,
    colframe=appcaseborder,
    colbacktitle=appcasetitle,
    coltitle=white,
    fonttitle=\bfseries\small,
    boxrule=0.7pt,
    arc=1.5mm,
    left=6pt,
    right=6pt,
    top=5pt,
    bottom=5pt,
  },
}
\newcommand{\appProblemCaseNeedspace}{\Needspace{0.48\textheight}}

\subsection{Synthetic Query Generation Prompt}
\label{app:query_prompt}

\begin{tcblisting}{appendixpromptbox,title={Prompt for Synthetic Query Generation}}
(*@\appPromptRole{\# system}@*)
You generate new scientific programming tasks for data expansion.
Rules:
- Use only the provided seed task as inspiration.
- Do not copy the seed task verbatim.
- Do not mention internal process details or prior instructions.
- Keep the new task in the same broad domain/task_type unless explicitly asked otherwise.
- Return one compact JSON object only. Do not include analysis, explanation, markdown, or code fences.
- The JSON fields must be exactly: problem_name, domain, subdomain, task_type, prompt, function_header, required_dependencies, tests.
- Keep prompt/function_header/tests concise enough that the full JSON fits in 2500 tokens.
- tests must be a list of 2 short Python assert snippets with concrete inputs, suitable for later verifier construction.

(*@\appPromptRole{\# user}@*)
Create one new synthetic scientific programming task from this seed.

Expansion target:
- seed_variant_index: {seed_idx}.
- preserve the same broad task_type and scientific coding flavor.
- vary constants, shapes, boundary cases, or scientific setup enough to avoid copying.
- produce a task that a 27B teacher can later solve and a verifier can check.

Output contract:
- Emit JSON only, starting with { and ending with }.
- Do not explain your reasoning.
- Keep tests short; each test string should be under 900 characters.

Seed task JSON:
{
  "problem_id": ...,
  "problem_name": ...,
  "domain": ...,
  "subdomain": ...,
  "task_type": ...,
  "problem_description": ...,
  "problem_io": ...,
  "required_dependencies": ...,
  "sub_steps": [
    {
      "step_number": ...,
      "step_prompt": ...,
      "function_header": ...
    }
  ]
}
\end{tcblisting}

\subsection{Failure-Mode Induction and Skill-Distillation Prompts}
\label{app:skill_distillation_prompts}

\begin{tcblisting}{appendixpromptbox,title={Failure-Mode Labeling Prompt}}
(*@\appPromptRole{\# system}@*)
You are an expert code-failure analyst for a scientific programming task.
For each failed sub-step, your job is to identify the MOST LIKELY root
cause(s) by contrasting the failed code with the successful sample (when given).

Output STRICT JSON, no markdown fences, no commentary, matching:
{
  "primary_labels": ["<label>", ...],
  "secondary_labels": ["<label>", ...],
  "novel_label": "<short_snake_case>|null",
  "evidence": "<one or two short sentences citing concrete code lines / variable names / shapes>",
  "confidence": <int 1-5>
}

Allowed labels (use EXACTLY these strings):
  - knowledge_gap_formula
  - knowledge_gap_algorithm
  - numerical_instability
  - shape_dtype_unit_error
  - api_signature_mismatch
  - logic_branch_error
  - spec_misread
  - error_propagation
  - thinking_exhaustion
  - import_or_dependency_error

Rules:
- Focus on the FAILED code's most likely root cause for THIS sub-step's failure.
- If the failed code is essentially empty or missing the target function definition, label it "thinking_exhaustion".
- If success exists and the diff is small (e.g., one line of array reshape), prefer "shape_dtype_unit_error" over "logic_branch_error".
- If no success sample is provided, you must still pick the most likely label(s) from the code + step description; lower the confidence.
- Cite EVIDENCE from the code (line snippets, variable names). Do not restate the problem.

(*@\appPromptRole{\# user}@*)
## Problem (id={problem_id}, name={problem_name})
domain={domain}  subdomain={subdomain}  task_type={task_type}
This is sub-step {step_idx_in_problem}/{tot_steps_in_problem}; step_id={step_id}.
Failure count over K=4 trials: {n_fail_K4}/4 (category={category}).

## Step description
{step_description}

## Step scientific background
{step_background_truncated_if_needed}

## Required function header
```python
{function_header}
```

## FAILED sample {i} (trial {trial}, judge={judge}) [{optional_flags}]
```python
{failed_code}
```

## SUCCESS sample {i} (trial {trial}) -- contrast against this
```python
{successful_code}
```

## Output
Return the JSON object only.
\end{tcblisting}

\begin{tcblisting}{appendixpromptbox,title={Failure-Family Induction Prompt}}
(*@\appPromptRole{\# system}@*)
You are a senior research engineer specializing in failure-mode taxonomy for LLM agents on
scientific computing&coding tasks. Your job: from a fixed batch of labeled failed sub-steps, induce a SMALL set of TASK-FAMILIES.

Project context:
- A "family" is defined by FAILURE AXIS, not by domain/subdomain.
- The goal is 1-3 procedural SOPs per family; SOPs that ONLY teach "what formula to use"
  don't generalize -- knowledge_gap_* failures should generally NOT be the attack target
  unless you can articulate a procedural meta-skill.
- p_partial bucket (P) = highest signal; p_none (N) bucket = often unteachable.
- Return EXACTLY 2 to 4 families.
- You assign each family a routing rule based on `primary_labels`: which primary labels
  belong to this family. Python will then route every step to a family by its primary label.
  Each primary label must appear in exactly ONE family's `routing_primary_labels`.

Output STRICT JSON, no markdown:
{
  "families": [
    {
      "family_id": "F1",
      "family_name": "<short snake_case>",
      "failure_axis": "<one sentence>",
      "definition": "<2-4 sentences: scope, what's IN and OUT>",
      "routing_primary_labels": ["..."],
      "representative_step_ids": ["...","..."],
      "cross_domain_evidence": "<one sentence>",
      "sop_attack_recommendation": "high" | "medium" | "low",
      "rationale_for_recommendation": "<one sentence>",
      "hypothesized_general_sop_seeds": ["<short procedural rule>", ...],
      "hypothesized_specific_knowledge_needs": ["<short note>", ...]
    }
  ],
  "global_notes": "<1-2 sentences>"
}

Constraints (will be validated):
- 2 <= len(families) <= 4
- routing_primary_labels across families form a PARTITION of these 10 labels:
  knowledge_gap_formula, knowledge_gap_algorithm, numerical_instability,
  shape_dtype_unit_error, api_signature_mismatch, logic_branch_error,
  spec_misread, error_propagation, thinking_exhaustion, import_or_dependency_error
- representative_step_ids must exist in the input.

(*@\appPromptRole{\# user}@*)
# Labeled failure steps ({n_steps} total)
Cols: step_id | cat(P=partial,N=none) | n_fail | task_type/domain | primary | secondary | evidence(short)

{step_id} | {P_or_N} | {n_fail_K4} | {task_type}/{domain} | {primary_labels} | {secondary_labels} | {short_evidence}
...

# Task
Induce 2-4 cross-domain failure families per system rules. Return JSON only.
\end{tcblisting}

\begin{tcblisting}{appendixpromptbox,title={Family-Level Skill Extraction Prompt}}
(*@\appPromptRole{\# system}@*)
You are extracting a FAMILY-LEVEL Standard Operating Procedure (SOP) for a coding
model on a scientific-programming task. The SOP will be injected into the prompt
prefix for all sub-steps that fall into this family.

HARD CONSTRAINTS (will be linted):
- Produce 1-3 SOPs (skills). Each skill has option-like schema + three utility
  dimensions + G/S section assignment.
- NO task-specific entities: no problem numbers, no step ids, no domain proper nouns
  (ewald, crank-nicolson, davidson, brownian, tweezer, jacobi, maxwell, etc.),
  no magic numbers tied to specific problems.
- Rules must be procedural decision-rules, NOT step-by-step recipes ("when you
  detect X, then Y" instead of "do A, then B, then C").
- "execution_policy" must contain BRANCHING decision points triggered by what the
  model observes; do NOT write a checklist of fixed steps.
- Less is more: 1-3 skills total, each compact (<=25 lines body).

For EACH skill, output JSON with these fields:
{
  "id": "<family_id>.<short_name>",
  "section": "G" | "S",
  "activation_condition": "<when to apply / when NOT to apply>",
  "execution_policy": "<decision-point text, multi-line OK>",
  "termination_condition": "<when this skill releases control to default policy>",
  "failure_mechanism": "<what concrete failure class this skill prevents>",
  "actionable_remedy": "<the concrete corrective action -- must be runnable>",
  "high_risk_blacklist": ["<anti-pattern>", ...],
  "evidence_step_ids": ["<step_id>", ...]
}

Return a JSON object: {"skills": [<skill>, ...], "rationale": "<one paragraph
explaining why these skills, why this G/S split, and what failures they target>"}

(*@\appPromptRole{\# user}@*)
# Family {family_id}: {family_name}
**Failure axis**: {failure_axis}
**Definition**: {definition}
**SOP attack value**: {sop_attack_recommendation}  ({rationale_for_recommendation})

**Seeds (from T8.4; you may keep, refine, drop, or replace)**:
- {hypothesized_general_sop_seed}

**Specific knowledge to keep external (do NOT bake into General SOPs)**:
- {hypothesized_specific_knowledge_need}

---
# Evidence (failed sub-steps + paired success contrast)
### step {step_id} ({task_type}, {n_fail_K4}/4 failed)
label: primary={primary_labels}  evidence: {label_evidence}
step description (truncated): {step_description}

FAILED code (trial {trial}, has_def={has_target_def}):
```python
{failed_code}
```

SUCCESS code (trial {trial}) -- contrast:
```python
{successful_code}
```

---
# Output
Produce 1-3 family-level SOPs per the system rules. Return STRICT JSON.
\end{tcblisting}

\subsection{Representative Generated Skills}
\label{app:skill_examples}

The promoted skill bundle contains family-level procedural skills rather than task-specific answers.
Below we show representative examples from the final bundle.

\begin{tcblisting}{appendixskillbox,title={F1.call\_audit: Specification and Interface Auditing}}
Activation: Apply when writing code that calls any function not defined within the current step's code block.

Execution policy:
- When you encounter a function call to an external/provided function:
  Is the function defined in this code block? If no, is it imported at the top of this block?
  If no, is it passed as a parameter from a prior step? If no, halt and add the required import or pass it as a parameter.
- When assigning the result of a function call, check whether the local variable name exactly matches the function name.
  If yes, rename the local variable immediately to prevent shadowing.
- When calling a multi-parameter function, verify argument count, order, and namespace qualification.

Avoid: assigning a call result to a variable with the same name as the called function; calling a function without verifying that it is imported or passed as a parameter; using bare names such as log, sqrt, or exp when a numpy namespace is expected.
\end{tcblisting}

\begin{tcblisting}{appendixskillbox,title={F2.shape\_broadcast\_audit: Array and Numerical Reasoning}}
Activation: Apply when writing code that creates, reshapes, combines, or reduces multi-dimensional arrays, especially when arrays of different rank are combined.

Execution policy:
- When two arrays will be combined, decide whether broadcasting aligns on the intended axes.
  If ambiguous, insert an explicit reshape or newaxis.
- When creating coordinate or index arrays from a dimension variable, verify that the coordinate length equals that dimension exactly.
- When using meshgrid, choose indexing='ij' when the first input maps to the first axis of the result.
- When performing axis-wise reduction, confirm that the axis argument collapses only dimensions that should collapse.
- After each critical array operation, insert a shape assertion or comment stating the expected output shape.

Avoid: combining arrays of different rank without explicit reshape; using meshgrid without specifying indexing mode; summing over axes without verifying which dimensions collapse.
\end{tcblisting}

\begin{tcblisting}{appendixskillbox,title={F3.conditional\_logic\_sanity: Conditional Logic and Dependency Flow}}
Activation: Apply when writing if/else branches, boolean masks, compound conditions, index-mapping expressions, or multi-target iteration logic.

Execution policy:
- For each conditional branch, enumerate the actual range of values the condition variable can take.
- For compound boolean expressions, wrap each sub-comparison in parentheses before combining with bitwise or logical operators.
- For index mapping or fancy indexing, write out the intended mapping as a concrete example and verify that row and column index arrays correspond to the intended source dimensions.
- For multi-element iteration, verify that each element receives its distinct, position-specific treatment rather than a duplicated operation.

Avoid: vacuous conditions, bare bitwise operators between comparisons, swapped row/column fancy-index mappings, and duplicated operators in multi-target iteration.
\end{tcblisting}

\section{Additional Experiment Diagnostics}
\label{app:exp_diagnostics}

\subsection{Data Construction Statistics}
\label{app:data_construction_stats}

Table~\ref{tab:app_data_construction} reports the retained SFT examples used by the two student training conditions.
The guided and control branches use the same source pool, candidate-generation budget, and construction protocol; their final retained sizes differ only because the same executable filtering stage accepts different numbers of generated solutions.
The synthetic examples are generated as new task instances and then completed by the 27B concretizer; they do not require benchmark-provided gold answers.
Executable checks are used only as filters when available.

\begin{table}[h]
\centering
\small
\caption{Retained SFT data after matched-budget construction, 27B solution generation, and executable filtering. ``Original'' denotes examples from the development-train task distribution; ``synthetic'' denotes newly generated query instances without pre-existing benchmark gold answers.}
\label{tab:app_data_construction}
\begin{tabular}{lrrrrl}
\toprule
\textbf{Training condition} & \textbf{Original} & \textbf{Synthetic} & \textbf{Total} & \textbf{Synthetic share} & \textbf{Teacher condition} \\
\midrule
Procedure-guided SFT & 395 & 1,700 & 2,095 & 81.1\% & Procedure-guided \\
No-procedure SFT control & 395 & 1,590 & 1,985 & 80.1\% & No procedure \\
\bottomrule
\end{tabular}
\end{table}

This construction-budget-matched comparison probes the contribution of \emph{procedure-guided} supervision beyond the more generic effect of 27B-generated solution SFT.
Both branches use the same development-train source pool, query-generation budget, and construction protocol; the intended treatment is whether the 27B model receives the consolidated procedures when producing code supervision.
The final pools contain 2,095 guided examples and 1,985 control examples because executable filtering retains 110 more solutions from the guided branch; this is a post-generation yield difference rather than an unmatched data-construction budget.

\subsection{Student Training Hyperparameters}
\label{app:training_hyperparameters}

We train both 9B students with full-parameter SFT for one epoch using a learning rate of $2\times10^{-6}$, a cosine schedule with a 0.03 warmup ratio, and AdamW with weight decay 0.1.
Training uses bfloat16 precision, a maximum sequence length of 49,152 tokens, one example per device, and four gradient-accumulation steps across eight GPUs, giving an effective global batch size of 32 examples.
We enable gradient checkpointing and DeepSpeed ZeRO Stage 2, and do not use sequence packing.
All runs use random seed 42 and the same optimization configuration for the procedure-guided and no-procedure students.

\subsection{Synthetic Data Composition}
\label{app:data_stats}

\begin{figure}[h]
\centering
\includegraphics[width=\linewidth]{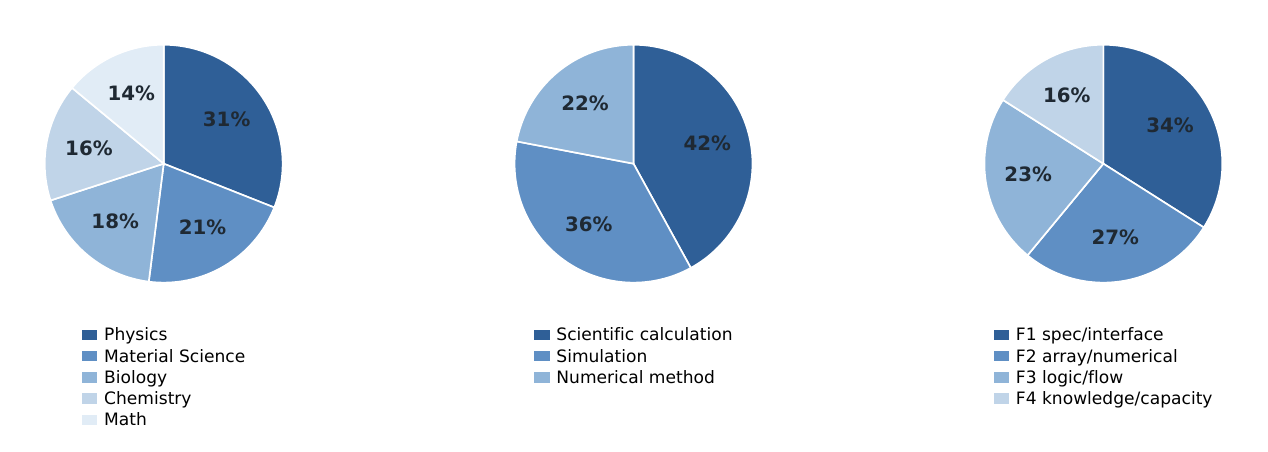}
\caption{Synthetic-data composition after stratified balancing. From left to right, the panels report broad-domain, task-type, and failure-family allocations used for data-construction control. The task-type panel includes all three SciCode categories, including numerical method. The distributions are intentionally not uniform: balancing is used to reduce source and category concentration while still reflecting seed availability, verifier yield, and the relative priority of different failure families.}
\label{fig:app_data_composition}
\end{figure}

The no-procedure control reuses the same query construction process but removes the procedure during 27B solution generation, providing a pipeline-matched comparison against SFT data construction without procedural guidance.

The net-gain view separates three effects.
First, the aggregate and split-level gains indicate that the procedure-guided student benefits beyond exposure to 27B-generated solutions alone, subject to the retained-size limitation above.
Second, the largest positive task-type bucket is simulation, while scientific calculation is positive but smaller.
Third, the broader numerical-method task bucket remains negative even though the narrower F2 failure-family diagnostic is positive, suggesting that shape/dtype supervision helps but does not yet cover the algorithmic numerical-method cases.

\subsection{Runtime Procedure Effects Across Model Groups}
\label{app:runtime_skill_table}

Table~\ref{tab:app_runtime_skill} reports the aggregate background-conditioned runtime-procedure deltas for all four model groups.
The pattern supports the main interpretation from Section~\ref{sec:capacity_mismatch}: runtime procedure injection is beneficial for the 27B model, but not for the 9B base model or the trained 9B students.
For the student models, the strongest deployment condition is procedure-free inference after SFT, which is the intended capability-consolidation setting.

\begin{table}[h]
\centering
\small
\caption{Runtime procedure effect under the background-conditioned aggregate evaluation. Results are sub-step/main-problem accuracy (\%).}
\label{tab:app_runtime_skill}
\begin{tabular}{lccc}
\toprule
\textbf{Model group} & \textbf{No procedure} & \textbf{With procedure} & \textbf{Procedure delta} \\
\midrule
9B base & 29.88 / 8.75 & 29.58 / 8.75 & -0.30 / +0.00 \\
27B teacher & 42.60 / 14.99 & 46.45 / 21.25 & +3.85 / +6.26 \\
No-procedure SFT & 31.61 / 13.75 & 30.74 / 10.75 & -0.87 / -3.00 \\
Procedure-guided SFT & 35.50 / 20.00 & 31.03 / 13.75 & -4.47 / -6.25 \\
\bottomrule
\end{tabular}
\end{table}

This table highlights a useful distinction between direct procedure use and capability consolidation.
The stronger model can use the runtime procedure directly, but the 9B student benefits most when the procedure has guided teacher solutions and the external artifact is removed at inference time.
In other words, the procedure is most useful as a data-generation condition, not as an additional prompt-time dependency for the smaller deployed model.

\subsection{Direct Procedure Consumption by Failure Family}
\label{app:direct_family_skill}

Table~\ref{tab:app_direct_family_skill} gives a development-side diagnostic over the labeled failure-family steps used by the procedure-synthesis pipeline.
The table should be read as a mechanism diagnostic over the same labeled family set rather than an independent benchmark result.

\begin{table}[h]
\centering
\small
\caption{Direct runtime procedure consumption over labeled development-train failure-family steps. Each cell reports sub-step accuracy without a runtime procedure $\rightarrow$ with a runtime procedure, followed by the gain. F4 is diagnostic and is not one of the promoted procedure families.}
\label{tab:app_direct_family_skill}
\begingroup
\setlength{\tabcolsep}{4pt}
\begin{tabular*}{\linewidth}{@{\extracolsep{\fill}}llccc@{}}
\toprule
\textbf{Family} & \textbf{Failure axis} & \textbf{\# steps} & \textbf{9B base} & \textbf{27B teacher} \\
\midrule
F1 & Specification/interface & 33 & 9.1 $\rightarrow$ 24.2 (+15.2) & 39.4 $\rightarrow$ 42.4 (+3.0) \\
F2 & Array/numerical reasoning & 15 & 26.7 $\rightarrow$ 20.0 (-6.7) & 40.0 $\rightarrow$ 33.3 (-6.7) \\
F3 & Logic/dependency flow & 11 & 27.3 $\rightarrow$ 27.3 (+0.0) & 54.5 $\rightarrow$ 63.6 (+9.1) \\
F4 & Knowledge/capacity gap & 62 & 6.5 $\rightarrow$ 6.5 (+0.0) & 22.6 $\rightarrow$ 33.9 (+11.3) \\
\bottomrule
\end{tabular*}
\endgroup
\end{table}

The family view refines the capacity-mismatch claim.
Some local interface checks can help the 9B model on F1, but the same model does not turn the procedure into reliable gains on logic/dependency flow, numerical robustness, or hard knowledge/capacity cases.
The 27B model is more able to convert the high-level procedure prompt into executable behavior on F3 and the diagnostic F4 bucket, while F2 remains difficult for both models.
This is the setting where teacher-mediated concretization is useful: the teacher can consume abstract procedural guidance on harder axes, and the student later learns from the resulting concrete trajectories.

\subsection{Diagnostic Takeaways from Fine-Grained Results}
\label{app:fine_diagnostic_takeaways}

The fine-grained results in Tables~\ref{tab:internalization_breakdown} and~\ref{tab:failure_family_breakdown} identify both where the current method works and where the next data synthesis round should focus.
The clearest positive buckets are simulation, Material Science, Math, Chemistry, and the F2/F3 failure families.
These buckets are consistent with the intended role of procedural knowledge: the 27B concretizer can convert high-level array checks, dependency-flow checks, and domain-specific coding patterns into concrete code supervision.

The negative bucket is also informative.
The broader numerical-method task type and the F1 specification/interface family remain weak relative to the no-procedure SFT control.
This does not contradict the capability-consolidation result; instead, it indicates that the current synthetic distribution under-represents the specific numerical-method failures that require more targeted supervision.
The next iteration should therefore allocate more balanced query synthesis budget to numerical-method seeds, interface-auditing failures, and verifier designs that stress formula-level and dependency-level boundary cases.

\section{Case Study}
\label{app:case_study}

\subsection{Case Studies of Rescued Failure Modes}
\label{app:case_studies}

Table~\ref{tab:case_studies} lists representative development-side cases where the 9B base model failed under the background-conditioned, procedure-free setting, while the procedure-guided SFT model succeeded under the same procedure-free inference condition.
The cases are selected from the paired comparison report used in Section~\ref{sec:skill_internalization_results}; all listed steps appear in the ``rescued'' set of that comparison.
F1--F3 correspond to the promoted procedural skill families, while F4 is included as a diagnostic family for knowledge or capacity limitations.

\begin{table}[h]
\centering
\small
\caption{Representative rescued cases by failure family. Each row is a SciCode sub-step where the 9B base failed and the procedure-guided SFT model succeeded under procedure-free inference.}
\label{tab:case_studies}
\begin{tabularx}{\linewidth}{p{0.09\linewidth}p{0.18\linewidth}p{0.22\linewidth}X}
\toprule
\textbf{Step} & \textbf{Failure family} & \textbf{Task} & \textbf{Observed base-model failure and relevant transferred behavior} \\
\midrule
65.1 & F1: spec/interface & GHZ protocol fidelity; implement a variadic tensor-product helper & Base failures defined \texttt{tensor()} without \texttt{*args}, causing references to \texttt{args} to be invalid. This matches the F1 interface-audit skill: verify the function signature and ensure every referenced argument is actually bound by the declared interface. \\
62.2 & F2: array/numerical & DMRG initial block construction; create single-site spin operators and block metadata & Base failures omitted explicit floating dtypes for spin matrices despite the specification requiring 2D arrays of floats. This matches the F2 array/dtype audit: annotate expected array types and verify that constructed arrays match the required numeric representation. \\
62.4 & F3: logic/dependency flow & DMRG block enlargement; update sparse block operators after adding a site & Base failures included undefined or mistyped live variables, such as using \texttt{Sm\_b} or assigning \texttt{conn\_Sp\_b\_sparse} from the wrong source variable. This matches the F3 dependency-flow skill: trace each computed value back to the live variable consumed downstream and verify branch/update consistency. \\
35.2 & F4: knowledge/capacity & Quantum-dot absorption spectrum; enumerate smallest quadratic combinations & Base failures used brute-force nested loops that enumerate indices rather than combination values, so early stopping can miss smaller later-index combinations. This F4 case is not evidence of a learned F4 procedural skill; it illustrates that teacher-mediated trajectories can still rescue some hard knowledge or algorithmic failures through concrete demonstrations. \\
\bottomrule
\end{tabularx}
\end{table}

These cases provide mechanistic examples consistent with the aggregate result: the teacher-mediated SFT model improves on failures that align with the procedural skills used during data construction, and it can also transfer concrete solution patterns for some diagnostic hard cases, even though the student does not receive those skills at inference time.

\subsection{Representative Mechanism Cases}
\label{app:mechanism_cases}

Table~\ref{tab:app_mechanism_cases} gives three compact cases corresponding to the main mechanisms discussed in the experiment section.
The cases are selected from the same paired diagnostics used for the main and appendix tables.

\begin{table}[h]
\centering
\small
\caption{Representative mechanism cases. The rows illustrate the abstraction--execution gap, procedure-guided supervision beyond the no-procedure SFT control, and a remaining weak numerical bucket.}
\label{tab:app_mechanism_cases}
\begin{tabularx}{\linewidth}{p{0.17\linewidth}p{0.18\linewidth}p{0.21\linewidth}X}
\toprule
\textbf{Mechanism} & \textbf{Step / family} & \textbf{Task} & \textbf{Diagnostic interpretation} \\
\midrule
Abstraction--execution gap & 14.1 / F3 & Brownian motion in an optical tweezer & Under runtime procedure injection, the 9B model fails while the 27B model succeeds. The failure pattern is a state-update bug: candidate solutions compute a new velocity but do not assign it back to the live state. This is exactly the kind of dependency-flow instruction that a stronger model can operationalize more reliably. \\
Procedure-guided supervision & 53.3 / F3 & Stochastic Lotka--Volterra simulation & The procedure-guided student is more reliable than the no-procedure SFT control on this dependency-flow case. The error pattern is an impossible frequency mask caused by misunderstanding the sign structure of FFT frequencies; the case is consistent with guided code supervision transferring concrete conditional-logic checks. \\
Residual weak bucket & Numerical method & Maxwell-equation and DMRG-style steps & The broader numerical-method bucket remains weaker than simulation and scientific calculation. Several hard cases require formula-level finite-difference signs, Hamiltonian construction, or solver-specific algorithmic details, which are not fully covered by the current shape/dtype-oriented F2 supervision. \\
\bottomrule
\end{tabularx}
\end{table}

\subsection{Procedure-Guided Gains over the No-Procedure Control}
\label{app:skill_vs_noskill_cases}

Table~\ref{tab:app_skill_vs_noskill_cases} lists representative sub-steps where the procedure-guided SFT student is more reliable than the no-procedure SFT control under the same background-conditioned, procedure-free inference setting.
These cases are not meant to be exhaustive.
They illustrate that the comparison in Section~\ref{sec:skill_internalization_results} is not merely a base-model improvement; procedure-guided code supervision adds behavior beyond what is obtained by no-procedure solution SFT alone.

\begin{table}[h]
\centering
\small
\caption{Representative sub-steps where procedure-guided SFT is more reliable than the no-procedure SFT control under procedure-free inference.}
\label{tab:app_skill_vs_noskill_cases}
\begin{tabularx}{\linewidth}{p{0.07\linewidth}p{0.18\linewidth}p{0.12\linewidth}p{0.12\linewidth}X}
\toprule
\textbf{Step} & \textbf{Task} & \textbf{Domain} & \textbf{Task type} & \textbf{Diagnostic interpretation} \\
\midrule
21.2 & GaAlAs absorption coefficient & Materials & Sci. calc. & A chained material-property computation where procedure-guided SFT is more stable than the no-procedure control; the case is consistent with the positive Material Science and scientific-calculation gains. \\
53.3 & Stochastic Lotka--Volterra & Biology & Simulation & A spectral-periodicity step with a frequency-mask logic error; it illustrates the F3-style conditional-logic benefit carried by procedure-guided supervision. \\
56.2 & Temporal niches & Biology & Simulation & A preference-ordering and growth-matrix construction step; the procedure-guided student is more reliable on this simulation case, matching the strongest task-type gain. \\
62.3 & DMRG block enlargement & Physics & Num. method & A Hamiltonian-construction step where procedure-guided supervision helps with concrete algorithmic scaffolding, showing that some numerical-method cases are recoverable despite the broader bucket weakness. \\
68.4 & Helium atom DMC & Chemistry & Simulation & A shape-rank indexing failure after norm reduction; this is a narrow F2-style array case that benefits from procedure-guided supervision, while broader numerical-method reasoning remains a separate weakness. \\
\bottomrule
\end{tabularx}
\end{table}

\subsection{Full Problem-Level Recovery Cases}
\label{app:problem_level_recovery}

The previous case tables focus on individual sub-steps.
We also include several complete task-level diagnostics to show what a capability-consolidation gain looks like when an entire step chain becomes executable.
The cases are selected from same-index paired runs in the background-conditioned, procedure-free deployment setting used in Section~\ref{sec:skill_internalization_results}: the original 9B base model is compared against the procedure-guided SFT student, and no runtime procedure is injected at inference time.
Together, they cover all three SciCode task types, include both multi-step chains and single-step full tasks, and span Math, Biology, Material Science, and Physics.

\appProblemCaseNeedspace
\begin{tcolorbox}[appendixcasebox,title={Problem 1: Conjugate gradient solver \hfill Math / numerical method}]
\textbf{Task.}
Solve a symmetric positive-definite linear system with the conjugate-gradient method, starting from an initial vector and stopping under a numerical tolerance.
Although this is a single-step problem, the step itself requires preserving the full iterative update structure: residuals, search directions, step sizes, convergence checks, and final return value.
\textbf{Primary axes:} F2 numerical iteration, residual/search-direction bookkeeping, and convergence control.

\vspace{3pt}
\begin{tabularx}{\linewidth}{p{0.12\linewidth}Xcc}
\toprule
\textbf{Step} & \textbf{Role in the task chain} & \makecell{\textbf{9B base}\\\textbf{before SFT}} & \makecell{\textbf{Procedure-guided SFT}\\\textbf{after SFT}} \\
\midrule
1.1 & Implement the complete conjugate-gradient solver with tolerance-based stopping. & \appFail & \appPass \\
\midrule
\textbf{Outcome} & \textbf{The single scored step determines main-problem correctness.} & \textcolor{red!65!black}{\textbf{not solved}} & \textcolor{green!45!black}{\textbf{solved}} \\
\bottomrule
\end{tabularx}

\vspace{4pt}
\textbf{Interpretation.}
The base model fails to assemble a numerically valid iterative solver.
After SFT, the student recovers the complete algorithmic scaffold, indicating that consolidated supervision can help even when the target behavior is not a local formula but an entire numerical routine.
\end{tcolorbox}

\appProblemCaseNeedspace
\begin{tcolorbox}[appendixcasebox,fontupper=\small,top=3pt,bottom=3pt,title={Problem 29: Gram--Schmidt orthogonalization \hfill Math / numerical method}]
\textbf{Task.}
Given an $N \times N$ NumPy array containing $N$ linearly independent vectors, implement a Gram--Schmidt procedure that returns normalized orthogonal vectors.
The task is decomposed into three scored steps: a vector normalization primitive, an inner-product primitive, and the final chained orthogonalization routine.
\textbf{Primary axes:} numerical-vector normalization, shape preservation, and chained composition.

\vspace{3pt}
\begin{tabularx}{\linewidth}{p{0.12\linewidth}Xcc}
\toprule
\textbf{Step} & \textbf{Role in the task chain} & \makecell{\textbf{9B base}\\\textbf{before SFT}} & \makecell{\textbf{Procedure-guided SFT}\\\textbf{after SFT}} \\
\midrule
29.1 & Normalize an input vector while preserving the NumPy array shape expected by later steps. & \appFail & \appPass \\
29.2 & Compute the inner product used by the projection terms in Gram--Schmidt. & \appPass & \appPass \\
29.3 & Compose the primitives into the full Gram--Schmidt orthogonalization routine. & \appPass & \appPass \\
\midrule
\textbf{Outcome} & \textbf{All scored steps must pass for the main problem to be solved.} & \textcolor{red!65!black}{\textbf{not solved}} & \textcolor{green!45!black}{\textbf{solved}} \\
\bottomrule
\end{tabularx}

\vspace{4pt}
\textbf{Interpretation.}
The base model fails the first primitive, so the task is not solved even though later isolated steps pass.
After procedure-guided SFT, the student fixes the primitive normalization step and keeps the downstream inner-product and orthogonalization steps correct, producing a contiguous all-pass chain.
This illustrates the intended deployment behavior: procedure-guided code supervision can be converted into procedure-free executable behavior in the student.
\end{tcolorbox}

\appProblemCaseNeedspace
\begin{tcolorbox}[appendixcasebox,title={Problem 53: Stochastic Lotka--Volterra \hfill Biology / simulation}]
\textbf{Task.}
Simulate a stochastic predator--prey system with Gillespie updates, evolve the trajectory, estimate oscillation periodicity, and compose the end-to-end solver.
This task stresses dependency flow across stochastic simulation primitives and helper reuse.
\textbf{Primary axes:} F3 dependency flow, stochastic branch logic, and helper composition.

\vspace{3pt}
\begin{tabularx}{\linewidth}{p{0.12\linewidth}Xcc}
\toprule
\textbf{Step} & \textbf{Role in the task chain} & \makecell{\textbf{9B base}\\\textbf{before SFT}} & \makecell{\textbf{Procedure-guided SFT}\\\textbf{after SFT}} \\
\midrule
53.1 & Implement one Gillespie event update with the correct cumulative propensity logic. & \appFail & \appPass \\
53.2 & Evolve prey and predator populations by repeatedly applying the event update. & \appFail & \appPass \\
53.3 & Estimate the dominant oscillation period from uneven stochastic samples. & \appPass & \appPass \\
53.4 & Compose the full predator--prey routine without dropping required helper behavior. & \appFail & \appPass \\
\midrule
\textbf{Outcome} & \textbf{All four scored steps must pass for the main problem to be solved.} & \textcolor{red!65!black}{\textbf{not solved}} & \textcolor{green!45!black}{\textbf{solved}} \\
\bottomrule
\end{tabularx}

\vspace{4pt}
\textbf{Interpretation.}
The base model breaks the chain at both local logic and helper-composition steps.
After procedure-guided SFT, the student keeps the stochastic update, trajectory evolution, periodicity analysis, and final composition simultaneously executable.
This is a representative F3-style recovery: the useful behavior is not a single formula, but preserving the dependency structure across the full simulation.
\end{tcolorbox}

\appProblemCaseNeedspace
\begin{tcolorbox}[appendixcasebox,title={Problem 56: Temporal niches \hfill Biology / simulation}]
\textbf{Task.}
Given species preferences, growth rates, and dilution, enumerate feasible depletion orders in a serial-dilution ecology model.
The task first filters logically possible resource-depletion orders, converts resource-level growth rates into temporal-niche growth rates, solves coexistence feasibility, and finally composes the full order-enumeration routine.
\textbf{Primary axes:} F2 array construction, F3 dependency flow, combinatorial filtering, and feasibility checking.

\vspace{3pt}
\begin{tabularx}{\linewidth}{p{0.12\linewidth}Xcc}
\toprule
\textbf{Step} & \textbf{Role in the task chain} & \makecell{\textbf{9B base}\\\textbf{before SFT}} & \makecell{\textbf{Procedure-guided SFT}\\\textbf{after SFT}} \\
\midrule
56.1 & Filter depletion orders that are inconsistent with the species preference constraints. & \appFail & \appPass \\
56.2 & Convert resource-level growth rates into growth rates over temporal niches. & \appPass & \appPass \\
56.3 & Solve the niche-duration feasibility system under a dilution factor. & \appPass & \appPass \\
56.4 & Compose filtering, conversion, and feasibility checks into the full enumerator. & \appFail & \appPass \\
\midrule
\textbf{Outcome} & \textbf{All four scored steps must pass for the main problem to be solved.} & \textcolor{red!65!black}{\textbf{not solved}} & \textcolor{green!45!black}{\textbf{solved}} \\
\bottomrule
\end{tabularx}

\vspace{4pt}
\textbf{Interpretation.}
The base model solves two intermediate computations but fails the logical-order filter and the final end-to-end composition.
The trained student preserves the full dependency chain, which is the behavior expected from procedure-guided supervision that repeatedly demonstrates array construction, constraint filtering, and helper reuse.
\end{tcolorbox}

\appProblemCaseNeedspace
\begin{tcolorbox}[appendixcasebox,title={Problem 51: Lennard--Jones molecular dynamics \hfill Material Science / simulation}]
\textbf{Task.}
Build a small molecular-dynamics pipeline: compute pairwise Lennard--Jones forces, aggregate net forces, perform a Velocity Verlet update, and run the resulting simulator.
The task combines scientific formula use with interface consistency across dependent functions.
\textbf{Primary axes:} formula scaffolding, F1 interface consistency, and helper dependency.

\vspace{3pt}
\begin{tabularx}{\linewidth}{p{0.12\linewidth}Xcc}
\toprule
\textbf{Step} & \textbf{Role in the task chain} & \makecell{\textbf{9B base}\\\textbf{before SFT}} & \makecell{\textbf{Procedure-guided SFT}\\\textbf{after SFT}} \\
\midrule
51.1 & Compute the pairwise Lennard--Jones force vector from a displacement vector. & \appFail & \appPass \\
51.2 & Aggregate all pairwise forces into a net force for each atom. & \appFail & \appPass \\
51.3 & Advance positions and velocities with one Velocity Verlet step. & \appFail & \appPass \\
51.4 & Compose repeated updates into a full simulation routine. & \appFail & \appPass \\
\midrule
\textbf{Outcome} & \textbf{All four scored steps must pass for the main problem to be solved.} & \textcolor{red!65!black}{\textbf{not solved}} & \textcolor{green!45!black}{\textbf{solved}} \\
\bottomrule
\end{tabularx}

\vspace{4pt}
\textbf{Interpretation.}
The base model fails at every layer of the chain, including the force formula, function-call interface, and downstream update composition.
The trained student produces a contiguous all-pass chain, suggesting that procedure-guided supervision can transfer both domain-specific formula scaffolding and F1/F3-style interface discipline.
\end{tcolorbox}

\appProblemCaseNeedspace
\begin{tcolorbox}[appendixcasebox,title={Problem 38: Reciprocal lattice vectors \hfill Physics / scientific calculation}]
\textbf{Task.}
Compute reciprocal lattice vectors from primitive lattice vectors, first through vector primitives and then through the final reciprocal-basis routine.
The task is a compact scientific-calculation chain where the final step requires preserving vector direction, dimensionality, and the $2\pi$ scaling convention.
\textbf{Primary axes:} formula convention, vector shape preservation, and dimensional consistency.

\vspace{3pt}
\begin{tabularx}{\linewidth}{p{0.12\linewidth}Xcc}
\toprule
\textbf{Step} & \textbf{Role in the task chain} & \makecell{\textbf{9B base}\\\textbf{before SFT}} & \makecell{\textbf{Procedure-guided SFT}\\\textbf{after SFT}} \\
\midrule
38.1 & Implement the cross-product primitive used by reciprocal-vector formulas. & \appPass & \appPass \\
38.2 & Compute 3D reciprocal vectors from three primitive vectors. & \appPass & \appPass \\
38.3 & Generalize the reciprocal-vector routine while preserving vector shape and formula convention. & \appFail & \appPass \\
\midrule
\textbf{Outcome} & \textbf{The final formula step is required for the main problem to be solved.} & \textcolor{red!65!black}{\textbf{not solved}} & \textcolor{green!45!black}{\textbf{solved}} \\
\bottomrule
\end{tabularx}

\vspace{4pt}
\textbf{Interpretation.}
The base model solves the local vector primitives but fails when the formula must be lifted into the final routine.
The SFT student preserves the required vector-valued reciprocal formula, illustrating recovery on a scientific-calculation case where shape and formula conventions are both active failure axes.
\end{tcolorbox}

\appProblemCaseNeedspace
\begin{tcolorbox}[appendixcasebox,title={Problem 6: Spatial filter I \hfill Physics / scientific calculation}]
\textbf{Task.}
Implement a Fourier-optics low-pass spatial filter that removes unwanted high-frequency components while preserving the central beam structure.
The threshold convention is part of the specification: the frequency exactly at the threshold should not be included in the pass mask.
\textbf{Primary axes:} F1 boundary convention, Fourier-domain mask construction, and shape-preserving inverse transform.

\vspace{3pt}
\begin{tabularx}{\linewidth}{p{0.12\linewidth}Xcc}
\toprule
\textbf{Step} & \textbf{Role in the task chain} & \makecell{\textbf{9B base}\\\textbf{before SFT}} & \makecell{\textbf{Procedure-guided SFT}\\\textbf{after SFT}} \\
\midrule
6.1 & Construct and apply the complete low-pass spatial filter with the required threshold convention. & \appFail & \appPass \\
\midrule
\textbf{Outcome} & \textbf{The single scored step determines main-problem correctness.} & \textcolor{red!65!black}{\textbf{not solved}} & \textcolor{green!45!black}{\textbf{solved}} \\
\bottomrule
\end{tabularx}

\vspace{4pt}
\textbf{Interpretation.}
The base model misses a specification-sensitive filtering behavior, where a small boundary convention changes the executable result.
After SFT, the student follows the mask construction and transform pipeline correctly, giving a compact scientific-calculation example of F1-style internalization.
\end{tcolorbox}

\appProblemCaseNeedspace
\begin{tcolorbox}[appendixcasebox,title={Problem 14: Brownian motion in an optical tweezer \hfill Physics / simulation}]
\textbf{Task.}
Simulate Langevin dynamics for a trapped microsphere with Mannella's leapfrog method and compute the mean-squared displacement over repeated trajectories.
The task is short, but both steps must remain correct for the whole simulation diagnostic to pass.
\textbf{Primary axes:} F3 state update, stochastic simulation flow, and specification reading.

\vspace{3pt}
\begin{tabularx}{\linewidth}{p{0.12\linewidth}Xcc}
\toprule
\textbf{Step} & \textbf{Role in the task chain} & \makecell{\textbf{9B base}\\\textbf{before SFT}} & \makecell{\textbf{Procedure-guided SFT}\\\textbf{after SFT}} \\
\midrule
14.1 & Update position and velocity with the stochastic leapfrog recurrence. & \appFail & \appPass \\
14.2 & Average repeated trajectories into a mean-squared-displacement estimate. & \appFail & \appPass \\
\midrule
\textbf{Outcome} & \textbf{Both scored steps must pass for the main problem to be solved.} & \textcolor{red!65!black}{\textbf{not solved}} & \textcolor{green!45!black}{\textbf{solved}} \\
\bottomrule
\end{tabularx}

\vspace{4pt}
\textbf{Interpretation.}
The base failure is a procedural simulation error: the computed new velocity is not reliably written back into the live state, and the downstream diagnostic inherits the mistake.
After SFT, the student executes the state update and the repeated-simulation wrapper together, matching the dependency-flow behavior targeted by the procedure-guided supervision.
\end{tcolorbox}

\end{document}